\documentclass{article} 
\usepackage{colm2024_conference}

\usepackage{microtype}
\usepackage{hyperref}
\usepackage{url}
\usepackage{booktabs}
\usepackage{graphicx}
\usepackage{amsmath}

\title{ComplexityNet: Increasing Language Model Inference \\ Efficiency by Learning Task Complexity}

\author{Henry Bae \& Kehang Zhu \thanks{ We give thanks to Professor H.T. Kung for providing helpful discussions } \\
Department of Physics\\
Harvard University\\
Cambridge, MA 02138, USA \\
henrybae@college.harvard.edu \\
kehang\_zhu@g.harvard.edu \\
\And
Aghyad Deeb \& Alex Fleury \\
Department of Computer Science\\
Harvard University\\
Cambridge, MA 02138, USA \\
\\
}

\colmfinalcopy 
\begin{document}

\maketitle

\begin{abstract}

We introduce ComplexityNet, a framework designed for the evaluation of task complexity and the allocation of tasks to Large Language Models (LLMs) of varying capabilities. This framework was applied to python code generation tasks, utilizing the Mostly Basic Python Problems (MBPP) dataset. To facilitate this, we first developed a set of labels to quantify task complexity accurately. 
Our methodology began with fine-tuning a small language model to predict the likelihood of generating accurate output across different LLMs. 
It achieved an accuracy of 79$\%$ in classifying task complexity, a significant increase from the 34$\%$ accuracy observed in the baseline model without fine-tuning. 
In the following step of allocating tasks, ComplexityNet reduced computational resource usage by 90$\%$ when compared to using the most complex model alone, while sustaining a high code generation accuracy of 86.7$\%$. 
This study demonstrates that fine-tuning smaller models to categorize tasks based on their complexity can lead to a more balanced trade-off between accuracy and efficiency in the use of Large Language Models. Our findings suggest a promising direction for optimizing LLM applications, especially in resource-constrained environments.

\end{abstract}

\section{Introduction}
The advent of large language models (LLMs) like GPT-4 signifies a remarkable advancement in the field of artificial intelligence \citep{OpenAI2023GPT4TR, Touvron2023Llama2O}, providing levels of natural language processing and generation that were previously unattainable.
However, this progress comes with its own set of challenges, particularly in the realm of inference costs. 
As these models become more advanced, the computational resources required to operate them increase significantly \citep{Li2020TrainBT}. For instance, it is speculated that GPT-4 utilizes around ten times more computations than its predecessor (roughly estimated by the API price), GPT-3.5 \citep{Brown2020LanguageMA}, highlighting a steep upward trend in resource demands. There is a uniform cost per token regardless of how difficult the task is.

This escalation in computational needs raises concerns about the sustainability and accessibility of these technologies\citep{Li2020TrainBT, Corro2023SkipDecodeAS}, especially for smaller entities or individual researchers. As more users gravitate towards these advanced models regardless the task simplicity, there's a significant waste of computational resources. While mainstream platforms offer manual model selection, allowing users to choose their preferred model \citep{OpenAI2023GPT4TR}, there's a lack of a systematic approach for automatically selecting the most appropriate model for a given task. 
Although we have LLM capability leaderboards for a wide range of tasks, the human brain struggles with switching between different models due to cognitive overload \cite{chandler1996cognitive, chen2011learners, nobre2024reading}.
This inefficiency becomes particularly apparent in tasks such as replicating code from documentation, where the output is extensive but the task's complexity does not align with the high computational expenditure. 
As a result, in the ongoing evolution of LLMs, it becomes critical to research and establish strategies to reduce these costs. Key questions arise, such as how to more effectively harness the capabilities of smaller models, and how to accurately assess task complexity to appropriately allocate tasks across various models. Addressing these considerations is vital not just for enhancing the financial viability of these models, but also for expanding their utility and accessibility. Resolving these concerns is an important step towards unlocking the full spectrum of possibilities offered by LLMs in a wide array of applications.



To resolve the challenges posed above, we introduce the following question: 

\begin{center}
\textbf{ For a given problem, what is the smallest model class that returns a correct answer?}
\end{center}

To quantify this question, we define the \textbf{complexity} of the problem as the simplest and the least capable LLM that is able to correctly accomplish the task given in the prompt. Smaller and less capable models will generate correct solutions for problems with low complexity but will fail to do so for problems with higher complexity. 

This notion of complexity can be difficult to define with respect to LLM's especially when the evaluation of the answer is subjective (For example: writing a meaningful poem). Therefore, we narrow the scope of problems to those that have definitive correct answers, such as mathematics problems (Ex: ``What is the sum of two numbers?'') or concrete programming problems (Ex: Function to generate an array of $n$ prime numbers). 

 To simplify the problem further, we restrict ourselves to three different language models with clear distinction in the number of parameters and their performance, Code Llama
 7B \citep{rozière2023code}, GPT-3.5, and GPT-4 \footnote{In this work, we use "GPT-4 Turbo" model. But we will simply denote it as "GPT-4".} \citep{OpenAI2023GPT4TR}. The table of models and their comparisons are shown below \footnote{We listed the data obtained at the time we ran this study -- December 1st, 2023}: 
\begin{table}[htbp]
\caption{Comparison of Language Models}
\begin{center}
\begin{tabular}{|c|c|c|}
\hline

\rule{0pt}{2ex}\textbf{Model}&\multicolumn{2}{|c|}{\textbf{Parameters and Inference Cost}} \\
\cline{2-3}
 \rule{0pt}{2.3ex} & \textbf{\textit{Number of Parameters}}& \textbf{\textit{Inference Cost}} \\
\hline
\rule{0pt}{2ex} Code Llama 7B & 7 Billion & \$0.0002/1K Token  \\
\hline
\rule{0pt}{2ex} GPT-3.5 & 175 Billion & \$0.002/1K Token \\
\hline
\rule{0pt}{2ex} GPT-4 & $>$1 Trillion* & \$0.03/1K Token \\
\hline
\end{tabular}
\label{tab1}
\end{center}
\end{table}

\begin{figure}[!ht]
\centering
\includegraphics[width=0.8\linewidth]{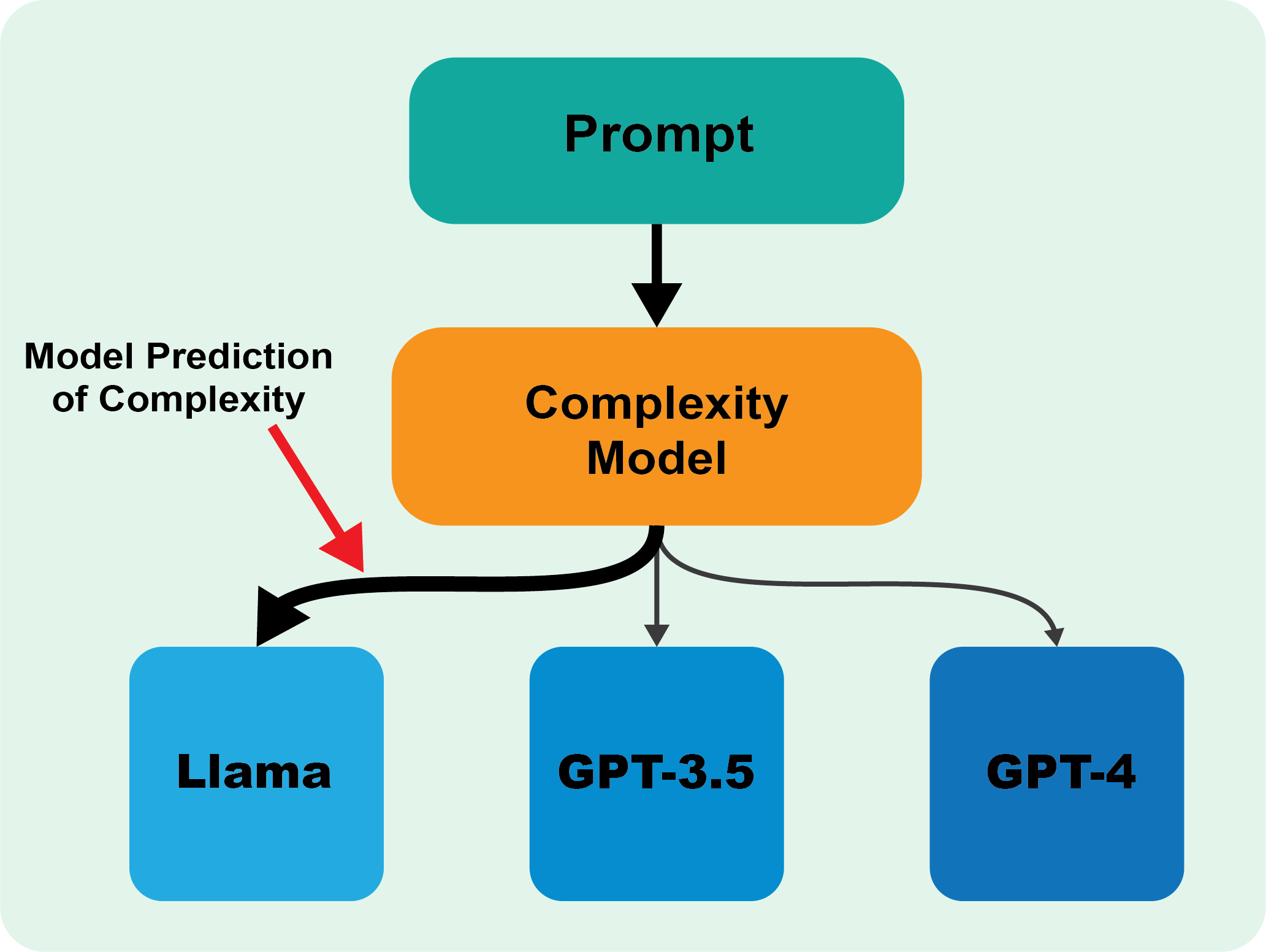}
\caption{Overview of the problem: The prompt is first fed through the complexity model then to one of the three models. We want to train a complexity model that picks the lowest cost-model that successfully accomplishes the task.}
\label{fig:problem_statement}
\end{figure}

Based on the three models, we aim to create a classification model that takes in the prompt as an input and outputs the complexity as we define it and choose the appropriate model based on the score. Figure~\ref{fig:problem_statement} illustrates this setup. 




\section{Related Works}

\textbf{Metrics for Determining the Capabilities of LLMs:}
The evaluation of LLMs has seen the development of diverse methodologies, each tailored to test specific capabilities. For instance, MMLU
 \citep{hendrycks2020measuring}
uses multiple-choice questions across a wide range of 57 subjects, from professional to academic, assessing the LLMs' understanding in these varied domains. Another approach, GSM8K
 \citep{cobbe2021training}, 
zeroes in on grade-school math problems to evaluate numerical reasoning. Similarly, MATH \citep{hendrycks2021measuring} 
challenges LLMs with math problems across five difficulty levels and seven sub-disciplines, providing a comprehensive metric for their mathematical capabilities. In addition, HumanEval
\citep{chen2022codet}
focuses on Python coding tasks to assess programming skills. Reading comprehension and arithmetic are evaluated in DROP
\citep{dua2019drop},
measured using the F1-score, and common-sense reasoning is tested in
\cite{zellers2019hellaswag} 
through multiple-choice questions.

These methodologies, while providing a percentage score for various tasks, do not provide a framework to analyze whether a model is capable of correctly completing a specific task, which is needed to efficiently utilize the model's abilities with minimal computational resources.

\textbf{LLM Autonomous Agents:}
The concept of autonomous agents in AI involves utilizing models that can independently manage complex tasks. 
Research in this area includes the HuggingGPT project and various studies advocating for the use of LLMs as controllers to manage existing AI models. For example, a paper from Microsoft Research \citep{shen2023hugginggpt} suggests using central agents to enhance multi-modal agent functionalities. Moreover, \cite{liu2023bolaa} propose using LLMs as a central controller to manage communication among multiple agents, each focusing on specific types of actions. Another significant contribution \cite{qin2023toolllm} involves fine-tuning LLaMA into ToolLLaMA, equipped with a neural API retriever, and its evaluation through ToolEval. 
\cite{manning2024automated} demonstrated the viability of using LLM agents not only as scientists for brainstorming ideas and designing experiments, but also as subjects for simulating social behaviors.

These advances point to a need for smarter model selection processes, moving beyond selection based on function descriptions, which often leads to inefficient computational resource usage.

\section{Methods}







\begin{figure}[!ht]
    \centering
    \includegraphics[width=0.9\linewidth]{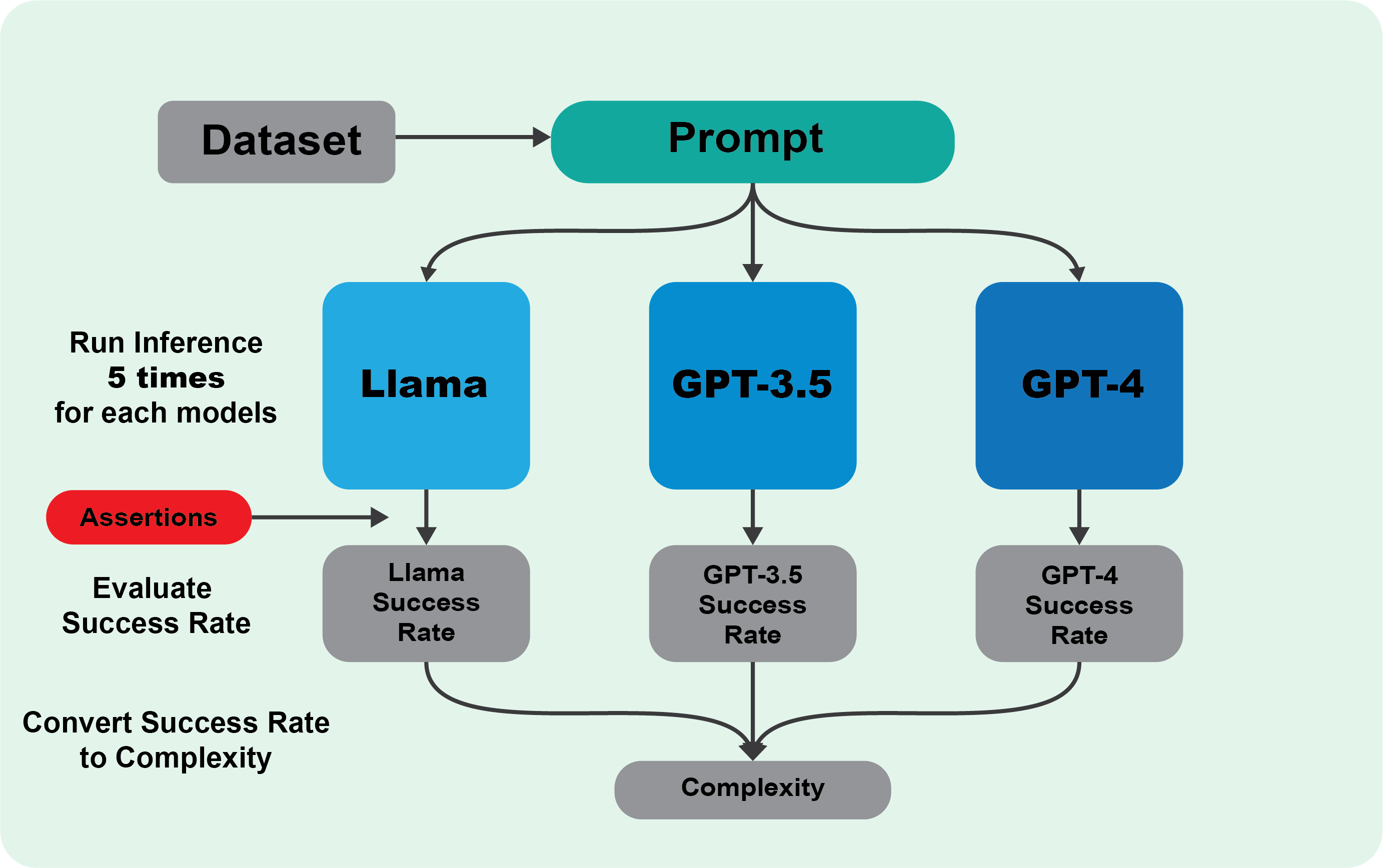}
    \caption{Overview of our approach. Each row of the dataset is fed through the three language models, and we store the success rate of each models. These success rates are used to generate a single complexity value for each prompts. }
    \label{Methodology}
\end{figure}

\subsection{Determining the task complexity for different LLMs}

Our approach first involves fine-tuning a small Language Model to output complexity levels based on the given task in prompt. To do so, we need to create a dataset with complexity labels for the prompts. These labels cannot be calculated manually, as the definition of complexity we presented here makes it entirely dependent on the LLM's outputs. Therefore, we adopt an empirical way to define the complexity level by the relative success rate for each task. Here, we only focus on tasks with clear answers and the success condition is simply that the output by LLM (taken after explanations) matches the solution. A visualization of our approach is depicted in Figure \ref{Methodology}.

Due to the stochastic nature of the outputs, we would query each of the $K$ primary LLMs \(L_1, L_2, ...L_K\) multiple times (M) for each task i (\(i = 1, ..., N\)) at a non-zero temperature. The number of successful trials will be denoted as \(X_{L_k, i}\), where $k \in [1, K]$ and $i\in [1, N]$ and $X_{L_k, i} \in [1, M]$. 
In the following, we will denote Code Llama, GPT-3.5 and GPT-4 as $L_1$, $L_2$ and $L_3$ respectively. 

We then classified a prompt's complexity into one of five classes, each represented by an integer $\in \{1, 2, 3, 4, 5\}$. 1 represents a very simple task and 5 represents a highly complex task. A task's complexity is determined through a mapping based on the results of the $K\times M$ total tests from before. 
\begin{figure*}
    \centering
    \includegraphics[width=\textwidth]{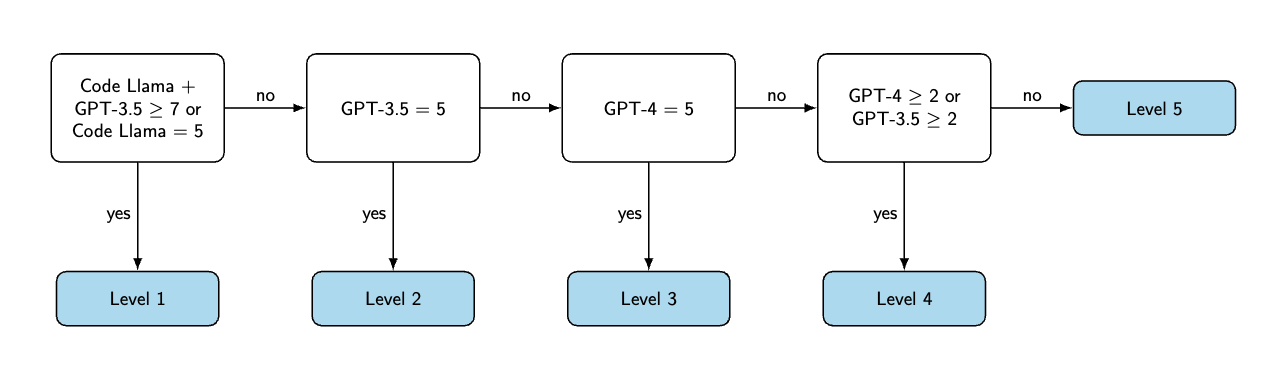}
    \caption{One example of the ordering mapping based on the success rate of each model at the task.}
    \label{Chart 1}
\end{figure*}



One example of the ordering mapping from the success rate of each LLM to the 5 complexity classes are defined in Figure~\ref{Chart 1} for task repetition time $M= 5$. For task $i$, level 1 corresponds to the case where $X_{L_1,i} =5$ or $X_{L_1,i} + X_{L_2,i}  \geq 7$, where we assume Code Llama was doing a good job. And level 2 maps to the case where the level 1 condition failed but $X_{L_2,i}  =5$ holds. Level 3 corresponds the scenario where the level 1$\&$2 conditions failed but $X_{L_3,i}  =5$ holds. If the level 1,2 and 3 conditions failed and the condition that $X_{L_2,i}  \geq 2$ or $X_{L_3,i}  \geq 2$ holds, we label these tasks to level 4.
If all of the conditions fails, we assign these tasks to level 5.

This mapping is based on the assumption that a model can reasonably solve a task that it solved correctly at least twice over five trials, factoring in higher confidence due to the ambiguity of the prompt, parameters, and verification conditions. We do not seek a \emph{guaranteed} correct answer (i.e. 5/5 correct solutions): we allocate the optimal model given the empirical results of the 15 total trials. But we have to note that more elaborated mappings can be applied.





\subsection{Automated pipeline for creating Task Complexity Dataset}

The method can in principle applied in any task class. For demonstration purpose, we chose to start with the most common use case -- Python coding task. 
And we picked up the Mostly Basic Python Problems (MBPP) dataset \citep{austin2021program}.
Each row of the dataset consists of a task that usually starts with, ``Write a function $\ldots$'', the Python code (solution) that accomplishes the task, and a series of Python assertion statements that can be used to verify the output of the model and get the success rate. The format and structure of the dataset allowed us to construct a single pipeline to run inference and verification.

To create the dataset, we constructed an automated pipeline to query the LLMs using tasks from the MBPP dataset.
We chose the set of LLMs to be GPT-4, GPT-3.5 and Code Llama with 7 billion parameters. The first two LLMs were called by OpenAI API and Code Llama was downloaded from Huggingface and ran locally.
We have also provided predetermined system prompt that would ensure the model is only outputting code without the explanation, and that it is following the same function definitions as the code that is in the assertion statement. The Code Llama seemed to perform poorly on the prompt we engineered for the two larger models so we cut down the prompt to only include the description of the function from the MBPP dataset and the format of the function that specifies the arguments.

The tasks were inputted to the three language models with a predetermined system prompt, and the output of the model was checked against the assertion statements from the dataset to verify the correctness of each LLM's answer. This was repeated 5 times (M=5) to reduce random noise in LLM's output and improve the robustness of the method.
We then adopted the mapping scheme as shown in Figure~\ref{Chart 1} to assign these tasks to different complexity levels.



Before fine-tuning, we also ``cleaned'' the dataset by removing all datapoints where each of Code Llama, GPT 3.5, and GPT 4 answered correctly $\frac{0}{5}$ times. We observed that in each of these cases, the models failed due to a mis-match between assertion code function definition and the prompt function definition (error in MBPP dataset). 






\subsection{Fine tuning process and task assignment}

Once this dataset is established, we can fine-tune a small language model to analyze the complexity of the prompts, which we're going to refer to as the Complexity model.
In this work, we chose the Complexity model as DaVinci-002 \citep{Brown2020LanguageMA}, which OpenAI API allows an easy fine-tuning.

And we split the whole complexity label dataset into a training set and a test set. Once the Complexity model is fine-tuned by the training set, it can assign out-of-sample tasks to different LLMs according to the predicted task complexity level.

In this study, we adopted a simple rule of assigning tasks to different LLMs by their complexity labels.
The scheme maps each of these complexity classes to one of Code Llama, GPT-3.5, or GPT-4. We assigned the complexity lavels 1 and 2 to Code Llama, Levels 3 and 4 to GPT-3.5, and Level 5 to GPT-4. Further studies are needed to improve the task level mapping and assignment rules.



\section{Results}



\subsection{Prediction accuracy on task complexity}

\begin{figure}[!ht]
\centering
\includegraphics[width=\linewidth]{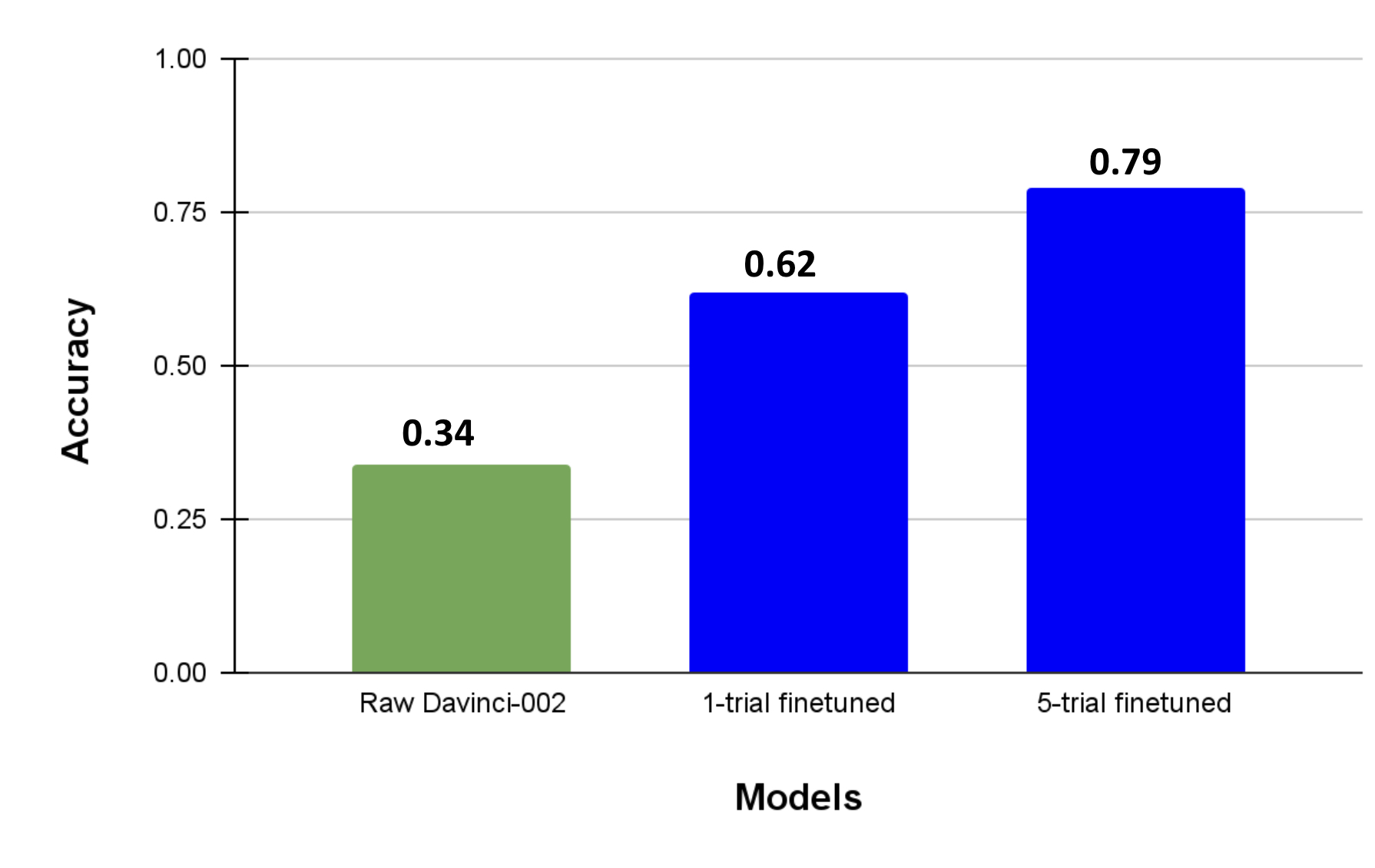}
\caption{
Comparison of the prediction accuracy of the the task complexity levels.
}
\label{fig:accuracy}
\end{figure}

The complexity dataset contains $N = 180$ of labels. We use $N_1 =144$ ($80\%$) for training and $N_2 =36$  ($20\%$) for testing. When running evaluations, we set the temperature of our LLMs to be 1. The  results indicated that a non-fine-tuned Davinci-002 achieved a $34\%$ accuracy. However, after fine-tuning, the model achieved $79\%$ accuracy, which yielded a significant 45$\%$ improvement. A comparison of the test set accuracy is shown in Figure \ref{fig:accuracy}.

Initially, we ran each task in MBPP dataset on each LLM for only one trial.
And we classified complexity into the classes 0, 1, and 2, which maps to Code Llama, GPT-3.5, and GPT-4, respectively. level 0 means Code Llama succeeded in the task and level 1 means the task can be done by GPT-3.5 and GPT-4 but not Code Llama. The highest level (level 2) means Code Llama and GPT-3.5 both failed but GPT-4 succeeded. The tasks that all of the three model failed will be dumped.
We followed the same procedure to fine-tune the Davinci-002 model and the accuracy is $62\%$. The non-fine-tuned model shows no significant change from the additional 4 trials, achieving a similar $32\%$ as before. So, 5-trial labeling method yield a 17$\%$ benefit over the single-trial labeling method.

This low accuracy reveals shortcomings of the single-trial labeling method. Providing an LLM one opportunity to solve a Python task does not effectively represent that model's capability to solve a problem of that complexity. We also observed inflated type II error rates that corroborated our skepticism.






 \subsection{Cost saving and Accuracy Trade-offs}

We now discuss the cost savings and the accuracy trade-offs associated with the utilization of the complexity model. To obtain a numerical estimate we need to make a series of assumptions. First, we approximate the cost of each models by observing their usage costs with API calls. From this, we can assign a unit cost 1 to Code Llama 7B model, a cost of 10 to GPT-3.5, and a cost of 100 to GPT-4. Second, we assume that the increase in computing for (five calls) to the complexity model is negligible considering the size of the output, a single token, as well as the complexity of the model that is in par with the least complex model. Lastly, we assume that we have an uniform distribution of dataset from different complexities. 

Below are cost savings estimates based on the empirical distribution of complexity observed in our $N = 180$ dataset and based on observed over and underestimations of complexity during inference. We benchmark our performances and measure savings on the assumption that users use GPT-4 for all of their tasks.

{\small
$$\text{Avg Compute of Correct Answer } (x) = 0.67 \cdot 1 + 10 \cdot 0.27 + 100 \cdot 0.06$$
$$\text{Avg Compute of Wrong Answer } (y) = 0.65 \cdot 1 + 10 \cdot 0.29 + 100 \cdot 0.06$$
$$\text{Compute Savings} = \frac{100 - (0.79 \cdot x + (1-0.79) \cdot y)}{100} = 0.90$$}
The dramatic 90\% decrease in savings is due to high accuracy of our complexity model and the ability for Code Llama and GPT-3.5 to succeed in python tasks. Notably, the method still sustained a high code generation accuracy of 86.7$\%$. 

We assume that the distribution of complexity in our dataset is reflective of the true distribution of difficulty of python problems in the MBPP dataset. We validated that the distribution of complexity in our test set during inference is approximately the same as the distribution in our entire dataset. We also note that the distribution of complexity labels 1, 2, and 3 are almost exactly equal in both the correct and incorrect answers. Both of these points corroborate our assumption.

\section{Discussion}

\subsection{Ecological niche of LLMs}
The high classification accuracy of the fine-tuned model with 5-Trial Labeling method and the resulting 90\% cost savings underscore a significant inefficiency in the prevailing approaches to LLM utilization, i.e., blindly using the most capable models for every tasks.
Our findings reveal the possibility of employing a top-down strategy for refining LLM inference processes that do not involve modifications at the model level.

Such findings echo the evolutionary study in machine behaviors \citep{rahwan2019machine}, raising the question of the ongoing relevance of smaller, older models in the face of advancements of larger and more capable models. However, akin to the diverse roles observed in natural ecosystems, where not only the largest or most advanced species thrive, these ``vintage'' LLMs may carve out unique niches. This suggests a dynamic, multi-faceted ecosystem of LLMs, where diversity rather than dominance dictates ecological balance.

\subsection{Extensibility of the Framework}
In this work, we analyzed the complexity of prompts that
involve coding tasks that correspond to a deterministic answer.
This opens questions about the feasibility of determining
complexity for more abstract tasks such as generation of essays
or poems. To do so, we can utilize different datasets such as the
MMLU\citep{hendryckstest2021}, a multi-task language understanding dataset benchmark, and determine whether our approach for determining
complexity. It is important to note that our overall procedure
can be generalized to any datasets that have assertions or
validation statements, and we speculate that training the model
on a combination of many datasets can help capture the
complexity of a wide range of tasks and prompts.


\section{Conclusion}
We presented a framework that applies a top-down optimization approach to enhance the inference efficiency of a set of LLMs.
A small language model was fine-tuned to determines a task's complexity and then match it with the best-suited, size-appropriate language model.
We implemented this approach in code generation tasks across three LLMs of different capabilities, achieving a remarkable $90\%$ reduction in inference costs while maintaining an accuracy rate of $86.7\%$. It's important to highlight that these results were achieved with substantial room for improvement in areas such as mapping, task assignment rules, and the fine-tuning process.
Next steps include exploring how this approach might work for a broader variety of tasks, especially those that are more complex or less defined than generating code.



\bibliography{colm2024_conference}
\bibliographystyle{colm2024_conference}






\end{document}